\documentclass{article}
\usepackage{spconf,amsmath,graphicx}

\title{Convolutional Neural Network Committees for Melanoma Classification with Classical And  Expert Knowledge Based Image Transforms Data Augmentation}
\twoauthors
{Cristina Nader Vasconcelos\sthanks{Thanks to NVidia for the donation of the Titan X GPU used in this research}}
{Departamento de Ci\^encia da Computa\c{c}\~{a}o\\
Instituto de Computa\c{c}\~{a}o \\
Universidade Federal Fluminense, Brazil}
{B\'{a}rbara Nader Vasconcelos}
{Servi\c{c}o de Dermatologia\\
Hospital Universit\'{a}rio Pedro Ernesto (Hupe)\\
Universidade Estadual do Rio de Janeiro, Brazil}

\begin{document}
%
\maketitle
\begin{abstract}
Skin cancer is a major public health problem, as is the most common type of cancer and represents more than half of cancer diagnoses worldwide.
Early detection influences the outcome of the disease and motivates our work.
We investigate the composition of CNN committees and data augmentation for the 
the ISBI 2017 Melanoma Classification Challenge (named Skin Lesion Analysis towards Melanoma Detection) facing the peculiarities of dealing with such a small, unbalanced, biological database.
For that, we explore committees of Convolutional Neural Networks trained over the ISBI challenge training dataset artificially augmented by both classical image processing transforms and image warping guided by specialist knowledge about the lesion axis and improve the final classifier invariance to common melanoma variations. 
\end{abstract}
\section{INTRODUCTION}
\label{sec:intro}


Skin cancer is the most common type of cancer and represents more than half of cancer diagnoses. There are two basic types of skin cancer named non-melanoma and melanoma, much rarer however, a much more serious disease.
%
Its incidence and overall mortality rates have been rising in recent decades \cite{Bita1},  therefore represents a substantial public health problem. 
Up to one fifth of the patients develop metastatic disease, which even may lead to death. However patients prognosis can be considered good when the melanoma is detected in the early stages.
Early detection and appropriate excision leads to cure rate of over 90\% in low risk melanoma patients. 
Innovative early detection programs, in combination with improved diagnostic tolls and new immunologic and molecular target treatments for advanced stages of the disease, may influence the outcome of the disease in the future \cite{Bita2}.

%

Dermatoscopy has proved valuable in visualizing the morphological structures in pigmented lesions for the diagnose of melanoma. It inspires the development of several Computer Vision Techniques for the Diagnosis of Skin Cancer based on dermoscopy images analysis \cite{Scharcanski_2013,Ammara_13}.
More recently, a great effort is being made as part of the ``International Skin Imaging Collaboration: Melanoma Project'' 
(ISIC \cite{ISIC}) to establish an archive annotated by experts to serve as a public resource of images for teaching and as a standard for the development of automated diagnostic systems. Currently, the ISIC Archive contains around 13,000 images associated with clinical metadata and expert annotations.

In 2016, the ISIC project organized the ISBI challenge \cite{GutmanCCHMMH16} in order push the development of dermoscopic image analysis tools for automated diagnosis of melanoma evaluated over a common standard. 
The ISBI challenge provides 900 annotated images as training data for participants to engage in all 3 components of lesion image analysis: lesion segmentation; detection and localization of features; and disease classification, which is the focus of this research. A separate test dataset containing 350 images is provided so that results of different methods can be compared.

Next, in 2017, a new challenge was organized so that a ternary problem is introduced: the classification between melanoma, nevus and seborrheic keratosis. A new database was published containing 
2000, where 374 from melanoma cases, 254 from seborrheic keratosis cases, and the remainder as benign nevi (1372). 
Validation and Tests were evaluated in new databases with respectively 150 and 600 images.

In this article we investigate the use of a Deep Learning approach, more specifically, based on Convolutional Neural Networks (CNNs) for the problem of Melanoma Classification trained on the ISBI challenge dataset.
CNNs are able to learn hierarchies of invariant features and classifiers from annotated datasets. 
They currently represent the state of the art solution for classification problems over natural images.
Since its 2012 edition, 
this is the case of the Imagenet Large Scale Visual Recognition annual Challenge (ILSVRC) \cite{ILSVRC15}   for object localization/detection and image/scene classification from images and videos at large scale.

The development of biomedical images classifiers (in our case, melanoma classification), have their own research peculiarities, being the most critical for the direct application of Deep Learning techniques, the considerably smaller annotated databases available,
as it is well known that CNNs usually require a large amount of training data in order to avoid overfitting. 
While the ISBI database used for training in this research contains 900 images, the ILSVRC 2012 participants received 10 million annotated images.  


\subsection{Hypotheses}
\label{subsec:Hypotheses}

We investigate how far a well known CNN topology (GoogLeNet \cite{GoogLenet14}) can be trained over the ISBI dataset, since it is expected that such a complex model suffers from severe overfitting when the amount of training samples received is two small. 

The following hypotheses are investigated:
%
%
(H1) it is possible to augment a CNN invariance to natural patterns from capturing conditions augmenting the training dataset with classical image processing operations, more specifically, exploring geometric and color transforms;
(H2) it is possible to augment a CNN invariance to biological patterns augmenting the training dataset with warped images preserving classes characteristics according to specialists knowledge (dermatologist); 
%
(H3) the CNN performance is affect by unbalanced training datasets (number of benign and malignant instances).


\subsection{Contributions}
\label{ssec:contrib}

Our methodology is based on a known CNN topology and investigates how such a small and unbalanced dataset can be extended using both classical and based on specialist knowledge image transforms. We believe that the contributions proposed can be applied to other topologies, such as for training residual neural networs, inproving the obtained result.
%

\section{RELATED}
\label{sec:related}
Data augmentation is being used by several classifiers based on learning processes in order to improve their solutions to natural invariances of the corresponding task.
In biological problems, CNNs are being used with relevant results. That is the case of the work of Ciresan et al \cite{CiresanGGS13} that obtained the first place at mitosis detection competition in 2012 Mitosis Detection in Breast Cancer Histological Images Challenge (ICPR 2012) with a 13-layer architecture. Similar to our work, they explored the creation of additional training instances by applying arbitrary rotations and/or mirroring.

%
%

Another widely investigated technique is the use of committees. In 2010, the MNIST handwriting recognition benchmark record dropped from $0.40\%$ to $0.27\%$ combinig seven CNNs \cite{CiresanMGS11}. 
Several other approaches combined the use of committees for smoothing the effects of non-optimal solutions with varying data augmentation approaches \cite{GoogLenet14,DBLP:conf/nips/2012,Residual}.
We focus on extending those ideas to the melanoma classification task, assuming the extra challenge of starting with a small annotated database.

\section{DATA AUGMENTATION}
\label{sec:dataAugm}
\subsection{Image Processing Techniques}
\label{ssec:classicalDataAugm}

Three transforms were included in order to augment the network invariance to geometric and color variations. 

\subsubsection{Geometric Augmentation}
The skin lesion scale and position within the image does not alter its final classification. 
Thus, input images were cropped to create new examples with the same label of the original one. 
The random cropping is expected to augment translation and scale invariances, but, different from cropping applying in training natural images classifiers,   
we enforce the assumption that the whole lesion is preserved in the new image produced, as its a requirement of the ISBI dataset and, more important, borders are used by specialist decisions criteria. 
Besides, the random crops produced maintained the original aspect ratio, as the lesions spatial proportions are also used by specialists in their melanoma/non-melanoma analysis. 
Besides, in order to augment the invariance to rotation and reflections, we 
replicated each image using eight unique combinations of rotation angles $\{0^o, 90^o, 180^o, 270^o\}$ and horizontal and vertical flips.

\subsubsection{Color Augmentation}

\begin{figure}[!t]
\footnotesize{
\begin{minipage}[b]{.31\linewidth}
  \centering
  \centerline{\includegraphics[width=2.0cm]{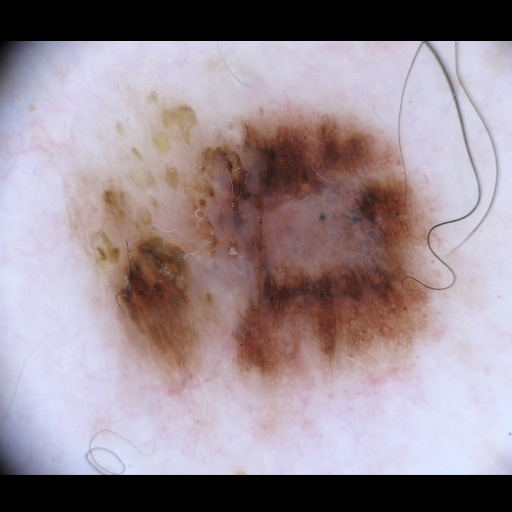}}
  \centerline{(a) Original}\medskip
\end{minipage}
\hfill
\begin{minipage}[b]{0.31\linewidth}
  \centering
  \centerline{\includegraphics[width=2.0cm]{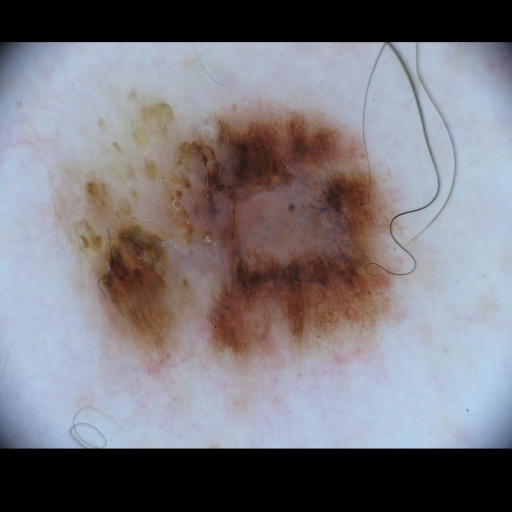}}
  \centerline{(b) Random PCA 1}\medskip
\end{minipage}
\hfill
\begin{minipage}[b]{.31\linewidth}
  \centering
  \centerline{\includegraphics[width=2.0cm]{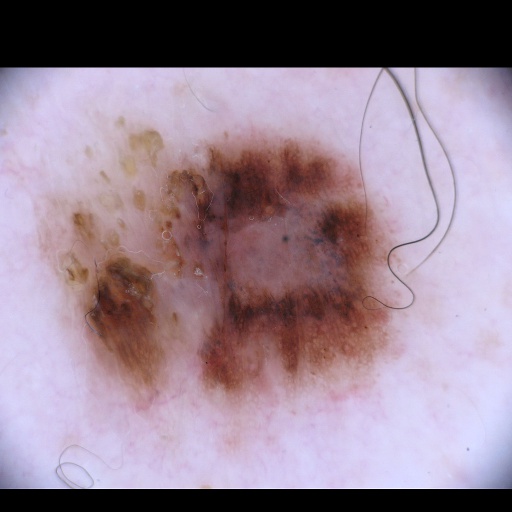}}
  \centerline{(c)  Random PCA 2}\medskip
\end{minipage}
\caption{Artificial samples created using PCA color transform}
\label{fig:colorAugmentation}
}
\end{figure}

Augmenting the color variety of an image dataset can be obtained by different techniques. However, in biological datasets, the colors created should correspond to plausible instances of the subject presented.  

In melanoma analysis we observe that the dermoscopic images present a systematic bias in their color distribution, occupying a very specific subset of the overall color spectrum. They correspond to skin and lesions biological variations, and also to the dermoscopic light color variations (the ISBI melanoma database contains two lighting distributions corresponding to a whitened and a bluish condition). 

In order to keep the artificial images produced  valid, we initially compute the Principal Component Analysis (PCA) of the colors from the training dataset
recovering its distribution main axes.
%
Artificial images are created adding multiples of the dataset principal components, with magnitudes proportional to their corresponding eigenvalues multiplied by a random factor obtained from a Gaussian distribution with mean $0.0$ and standard deviation $\sigma=0.2$. 
The adopted value of $\sigma$ was found creating samples with increasing magnitude and validating its appearance with a specialist. This validation process showed that higher values of delta let to unnatural images. Samples created are illustrated in Figure \ref{fig:colorAugmentation}.


\subsection{Data Augmentation based on Specialists Knowledge}
\label{ssec:Specialists}

\begin{figure}[t!]
\footnotesize{
\begin{minipage}[b]{.24\linewidth}
\centering
\centerline{\includegraphics[width=2.0cm,height=2.0cm]{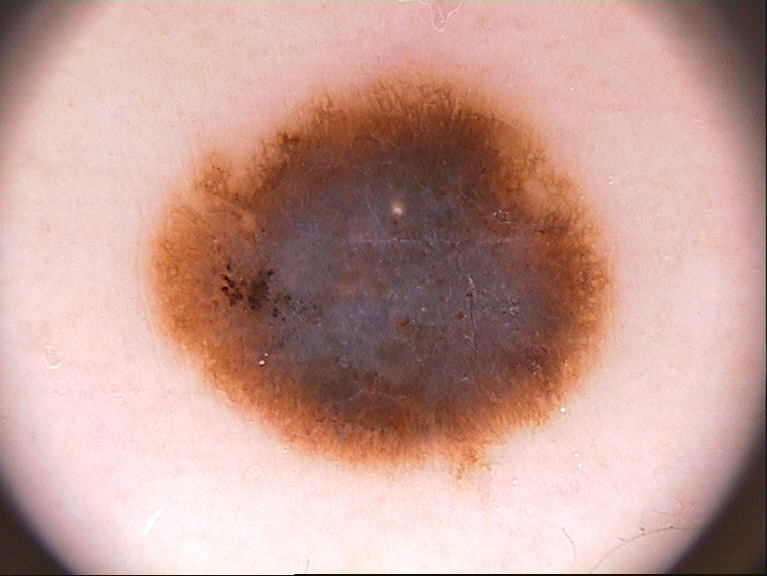}}
  \centerline{(a) Original}\medskip
\end{minipage}
\begin{minipage}[b]{0.24\linewidth}
  \centering
 \centerline{\includegraphics[width=2.0cm,height=2.0cm]{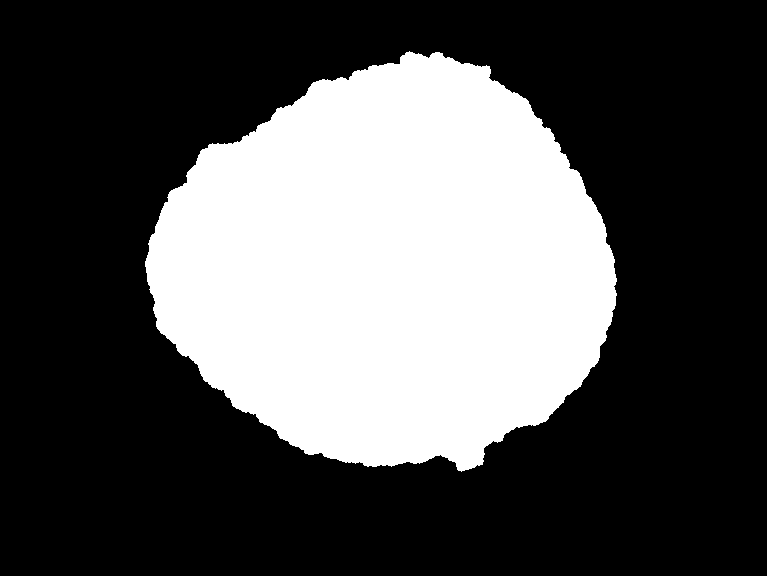}}
  \centerline{(b) Segmentation}\medskip
\end{minipage}
\begin{minipage}[b]{.24\linewidth}
  \centering
 \centerline{\includegraphics[width=2.0cm]{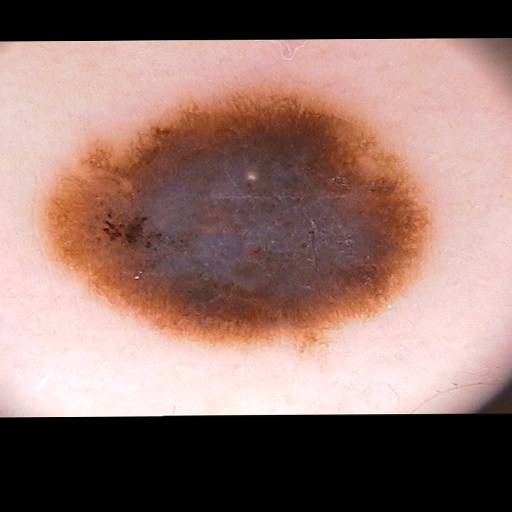}}
  \centerline{(c) Artificial 1}\medskip
\end{minipage}
\begin{minipage}[b]{0.24\linewidth}
  \centering \centerline{\includegraphics[width=2.0cm]{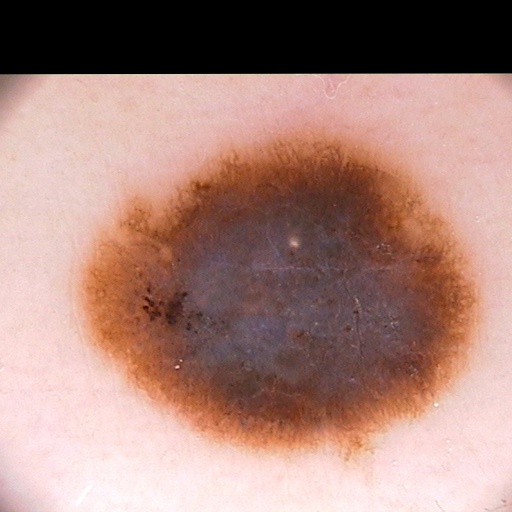}}
  \centerline{(d) Artificial 2}\medskip
\end{minipage}
\caption{Distortion of the original image by lesion axis analysis}
\label{fig:thin}
}
\end{figure}

While the lack of dependence on prior knowledge and human effort in designing features is a major advantage for CNNs, we argue that specialist knowledge can be used in order to expand the training samples available, while preserving the annotation.
The melanoma specialist analysis is made over the observation of the symmetries and patterns around the lesion main orthogonal axis. 
Those symmetries (or anti-symmetries) should be preserved in the new artificial images as they imply the classification of the lesion.

We proposed a data augmentation scheme based on the distortion of the lesion main axis size, while maintaining the patters and symmetries of the lesion. For creating such artificial image, we approximate the lesion segmentation (Figure \ref{fig:thin} (b)) with an ellipse and use its the main axis as controllers of our distortion. The size of the axis is randomly increased or reduced, while the axis directions are maintained. 

The warped image is created with a thin plate spline mapping \cite{Sibson_1991} between five points of the original image into new positions of the artificial image. They are the center of the ellipse (controlling central position variations), and the four axis end points, that are randomly translated in $20\%$ of their original axis sizes (positive or negative size changes), but keeping their original directions (Figure \ref{fig:thin}).

\section{CONVOLUTIONAL NEURAL NETWORKS: GoogLeNet}
\label{sec:CNNS}

\begin{figure}[t!]
 \centering
 \includegraphics[width=0.6\columnwidth]{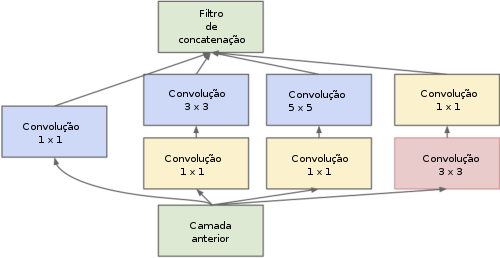}
 \caption{Inception module \cite{GoogLenet14}: mixing filters of different sizes.
 %
 }
\label{fig:inception}
\end{figure}

The goal of this paper is not to develop a new CNN architecture, but to evaluate how far solutions based on CNN can be applied to the Melanoma Classification problem, given the ISBI dataset size restrictions.
A well known topology was chosen for implementing the proposed classifiers, 
named GoogLeNet~\cite{GoogLenet14},
in order to clarify our proposal contributions in a replicable framework. 
The GoogLeNet is the ILSVRC 2014 winner from Szegedy et al. from Google ~\cite{GoogLenet14}. 
Its main contribution was the development of the Inception Module (Figure \ref{fig:inception}) that dramatically reduced the number of parameters in the network, creating a deeper and wider topology. 

The input signal that enters an inception module is processed by filters of different sizes, while in previous networks the convolution kernels were typically uniform.
%
The topology named GoogLeNet is formed with
Inception modules stacked on top of each other, leading to a 22-layer deep model.

\section{Experiments}
\label{ssec:Expirements}

All the experiments presented adopted the same CNN topology and hyper-parameters (from GoogLeNet \cite{GoogLenet14}) in order to focus
exclusively on the improvements obtained by the data augmentations proposed. They are: 10 epochs; a fixed learning rate of $10^{-4}; and a batch of 256 images.$


The experiments started with a random split of the ISBI training set (that contains 2000 images) into five folders of 400 images each, maintaining the original distribution of $18.7\%$ melanoma, $12.7\%$ nevus, $68.6\%$ seborrheic keratosis  within each created folder ($\approx$0.24). 
A 5-fold cross validation was applied so that images from four folders were used in each training, while a fifth as a validation set. 
The data augmentations were only applied to those folders considered training folders.

Tests  over unseen images were made against the ISBI 2017 test and validation datasets, but final tests results are not available at the time of writing.




The results obtained where obtained starting the training procedures with the GoogLeNet \cite{GoogLenet14} ILSVC 2014 \cite{ILSVRC15} public set of weights pre-calculated over a larger dataset of natural images. Such a fine-tuning reduced the overfitting and training time observed in all experiments as expected.

\begin{table}[!ht]

\caption{Results of the proposed committees composed with GoogLeNets and different data augmentation schemes investigated versus ISBI challenge first and second best results}
\label{tab:resultsCommittees} 
\centering
\footnotesize{
\begin{tabular}{c|c}
\multicolumn{1}{c}{Team name} 
& \multicolumn{1}{c}{Score} \\
   \hline
Kazuhisa Matsunaga	&	0.958 \\
RECOD Titans	&	0.951 \\
T D	&	0.943 \\
Ours & 	0.932
\end{tabular}
}
\end{table}

\section{CONCLUSIONS}
\label{sec:onclusions}

Melanoma early detection programs, in combination with improved diagnostic tolls are crucial for the outcome of the disease. Besides, deep learning approaches are conquering the state of the art in several image classification tasks. 

We explored how a deep learning solution can be extended for dealing with a small and unbalanced biomedical dataset.

The data augmentations proposed, together with the balanced of the created artificial datasets and the proposed committees groping classifiers of CNNs trained over balanced sets obtained the a result that was only $2.6\%$ bellow the best result published using the validation dataset, validating the hypotheses presented in Section \ref{subsec:Hypotheses} 
investigated by this work. Such a small difference can be reduced using a stronger CNN topology, such as a Residual CNN \cite{Residual}, motivating further investigation of our approach.

Finally, we note that further comparisons with the methods proposed by participants of the ISBI melanoma challenge were not made by the lack of public descriptions of their approaches at the time of writing. 

%

\bibliographystyle{IEEEbib}
\bibliography{refs}

\end{document}